\documentclass[10pt,twocolumn,letterpaper]{article}

\usepackage{cvpr}
\usepackage{times}
\usepackage{epsfig}
\usepackage{graphicx}
\usepackage{amsmath}
\usepackage{amssymb}
\usepackage{multirow}
\usepackage{color,soul}
\usepackage[table,xcdraw]{xcolor}

\usepackage[pagebackref=true,breaklinks=true,letterpaper=true,colorlinks,bookmarks=false]{hyperref}

\cvprfinalcopy 

\def\cvprPaperID{****} 

\ifcvprfinal\fi
\begin{document}

\title{Auditing ImageNet: Towards a Model-driven Framework for Annotating Demographic Attributes of Large-Scale Image Datasets}

\author{
    Chris Dulhanty\\
    University of Waterloo\\
    Waterloo, Ontario, Canada\\
    {\tt\small chris.dulhanty@uwaterloo.ca}
    \and
    Alexander Wong\\
    University of Waterloo\\
    Waterloo, Ontario, Canada\\
    {\tt\small a28wong@uwaterloo.ca}
}

\maketitle

\begin{abstract}
   The ImageNet dataset ushered in a flood of academic and industry interest in deep learning for computer vision applications. Despite its significant impact, there has not been a comprehensive investigation into the demographic attributes of images contained within the dataset. Such a study could lead to new insights on inherent biases within ImageNet, particularly important given it is frequently used to pretrain models for a wide variety of computer vision tasks. In this work, we introduce a model-driven framework for the automatic annotation of apparent age and gender attributes in large-scale image datasets. Using this framework, we conduct the first demographic audit of the 2012 ImageNet Large Scale Visual Recognition Challenge (ILSVRC) subset of ImageNet and the `person' hierarchical category of ImageNet. We find that 41.62\% of faces in ILSVRC appear as female, 1.71\% appear as individuals above the age of 60, and males aged 15 to 29 account for the largest subgroup with 27.11\%. We note that the presented model-driven framework is not fair for all intersectional groups, so annotation are subject to bias. We present this work as the starting point for future development of unbiased annotation models and for the study of downstream effects of imbalances in the demographics of ImageNet. Code and annotations are available at: \url{http://bit.ly/ImageNetDemoAudit}
\end{abstract}

\section{Introduction}

ImageNet \cite{deng2009imagenet}, released in 2009, is a canonical dataset in computer vision. ImageNet follows the WordNet lexical database of English \cite{miller1995wordnet}, which groups words into \textit{synsets}, each expressing a distinct concept. ImageNet contains 14,197,122 images in 21,841 synsets, collected through a comprehensive web-based search and annotated with Amazon Mechanical Turk (AMT) \cite{deng2009imagenet}. The ImageNet Large Scale Visual Recognition Challenge (ILSVRC) \cite{russakovsky2015imagenet}, held annually from 2010 to 2017, was the catalyst for an explosion of academic and industry interest in deep learning. A subset of 1,000 synsets were used in the ILSVRC classification task. Seminal work by Krizhevsky \textit{et al.} \cite{krizhevsky2012imagenet} in the 2012 event cemented the deep convolutional neural network (CNN) as the preeminent model in computer vision.

Today, work in computer vision largely follows a standard process: a pretrained CNN is downloaded with weights initialized to those trained on the 2012 ILSVRC subset of ImageNet, the network is adjusted to fit the desired task, and transfer learning is performed, where the CNN uses the pretrained weights as a starting point for training new data on the new task. The use of pretrained CNNs is instrumental in applications as varied as instance segmentation \cite{he2017mask} and chest radiograph diagnosis \cite{rajpurkar2018deep}. 

By convention, computer vision practitioners have effectively abstracted away the details of ImageNet. While this has proved successful in practical applications, there is merit in taking a step back and scrutinizing common practices. In the ten years following the release of ImageNet, there has not been a comprehensive study into the composition of images in the classes it contains.

This lack of scrutiny into ImageNet's contents is concerning. Without a conscious effort to incorporate diversity in data collection, \textit{undesirable biases} can collect and propagate. These biases can manifest in the form of patterns learned from data that are influential in the decision of a model, but are not aligned with values of society \cite{stock2018convnets}. Age, gender and racial biases have been exposed in word embeddings \cite{bolukbasi2016man}, image captioning models \cite{anne2018women}, and commercial computer vision gender classifiers \cite{buolamwini2018gender}. In the case of ImageNet, there is some evidence that CNNs pretrained on its data may also encode undesirable biases. Using adversarial examples as a form of model criticism, Stock and Cisse \cite{stock2018convnets} discovered that prototypical examples of the synset `\textit{basketball}' contain images of black persons, despite a relative balance of race in the class. They hypothesized that an under-representation of black persons in other classes may lead to a biased representation of `\textit{basketball}'.

This paper is the first in a series of works to build a framework for the audit of the demographic attributes of ImageNet and other large image datasets. The main contributions of this work include the introduction of a model-driven demographic annotation pipeline for apparent age and gender, analysis of said annotation models and the presentation of annotations for each image in the training set of the ILSVRC 2012 subset of ImageNet (1.28M images) and the `\textit{person}' hierarchical synset of ImageNet (1.18M images).

\begin{table}
\begin{center}
\footnotesize
\begin{tabular}{|
>{}c |
>{}c |
>{}c |
>{}c |
>{}c |}
\hline
\multicolumn{2}{|c|}{{\color[HTML]{000000}}} & \multicolumn{3}{c|}{{\color[HTML]{000000} \textbf{Gender}}} \\ \cline{3-5} 
\multicolumn{2}{|c|}{\multirow{-2}{*}{{\color[HTML]{000000} \textbf{Average Precision (\%) }}}} & {\color[HTML]{000000} \textbf{Male}} & {\color[HTML]{000000} \textbf{Female}} & {\color[HTML]{000000} \textbf{All}} \\ \hline
{\color[HTML]{000000} } & {\color[HTML]{000000} \textbf{0-14}} & {\color[HTML]{000000} 93.75} & {\color[HTML]{000000} 100.00} & {\color[HTML]{000000} 97.44} \\ \cline{2-5} 
{\color[HTML]{000000} } & {\color[HTML]{000000} \textbf{15-29}} & {\color[HTML]{000000} 99.20} & {\color[HTML]{000000} 100.00} & {\color[HTML]{000000} 99.61} \\ \cline{2-5} 
{\color[HTML]{000000} } & {\color[HTML]{000000} \textbf{30-44}} & {\color[HTML]{000000} 100.00} & {\color[HTML]{000000} 99.11} & {\color[HTML]{000000} 99.76} \\ \cline{2-5} 
{\color[HTML]{000000} } & {\color[HTML]{000000} \textbf{45-59}} & {\color[HTML]{000000} 100.00} & {\color[HTML]{000000} 100.00} & {\color[HTML]{000000} 100.00} \\ \cline{2-5} 
{\color[HTML]{000000} } & {\color[HTML]{000000} \textbf{60+}} & {\color[HTML]{000000} 99.17} & {\color[HTML]{000000} 100.00} & {\color[HTML]{000000} 99.24} \\ \cline{2-5} 
\multirow{-7}{*}{{\color[HTML]{000000} \textbf{\begin{tabular}[c]{@{}c@{}}Age\\ Group\end{tabular}}}} & {\color[HTML]{000000} \textbf{All}} & {\color[HTML]{000000} 99.48} & {\color[HTML]{000000} 99.74} & {\color[HTML]{000000} 99.54} \\ \hline
\end{tabular}
\end{center}
\caption{Face detection model average precision on a subset of FDDB, hand-annotated by the author for apparent age and gender}
\label{table:fddb-face-detection}
\end{table}

\begin{table}
\begin{center}
\footnotesize
\begin{tabular}{|
>{}c |
>{}c |
>{}c |
>{}c |
>{}c |}
\hline
\multicolumn{2}{|c|}{{\color[HTML]{000000} }} & \multicolumn{3}{c|}{{\color[HTML]{000000} \textbf{Gender}}} \\ \cline{3-5} 
\multicolumn{2}{|c|}{\multirow{-2}{*}{{\color[HTML]{000000} \textbf{Mean Average Error (years)}}}} & {\color[HTML]{000000} \textbf{Male}} & {\color[HTML]{000000} \textbf{Female}} & {\color[HTML]{000000} \textbf{All}} \\ \hline
{\color[HTML]{000000} } & {\color[HTML]{000000} \textbf{0-14}} & {\color[HTML]{000000} 8.26} & {\color[HTML]{000000} 9.37} & {\color[HTML]{000000} 8.77} \\ \cline{2-5} 
{\color[HTML]{000000} } & {\color[HTML]{000000} \textbf{15-29}} & {\color[HTML]{000000} 3.24} & {\color[HTML]{000000} 3.65} & {\color[HTML]{000000} 3.57} \\ \cline{2-5}
{\color[HTML]{000000} } & {\color[HTML]{000000} \textbf{30-44}} & {\color[HTML]{000000} 4.52} & {\color[HTML]{000000} 4.93} & {\color[HTML]{000000} 4.72} \\ \cline{2-5} 
{\color[HTML]{000000} } & {\color[HTML]{000000} \textbf{45-59}} & {\color[HTML]{000000} 4.43} & {\color[HTML]{000000} 5.50} & {\color[HTML]{000000} 4.81} \\ \cline{2-5} 
{\color[HTML]{000000} } & {\color[HTML]{000000} \textbf{60+}} & {\color[HTML]{000000} 7.70} & {\color[HTML]{000000} 9.48} & {\color[HTML]{000000} 8.40} \\ \cline{2-5} 
\multirow{-7}{*}{{\color[HTML]{000000} \textbf{\begin{tabular}[c]{@{}c@{}}Age\\ Group\end{tabular}}}} & {\color[HTML]{000000} \textbf{All}} & {\color[HTML]{000000} 5.13} & {\color[HTML]{000000} 5.29} & {\color[HTML]{000000} 5.22} \\ \hline
\end{tabular}
\end{center}
\caption{Apparent age model mean average error in years on APPA-REAL test set}
\label{table:appa-real-age}
\end{table}

\section{Diversity Considerations in ImageNet} 
Before proceeding with annotation, there is merit in contextualizing this study with a look at the methodology proposed by Deng \textit{et al.} in the construction of ImageNet. A close reading of their data collection and quality assurance processes demonstrates that the conscious inclusion of demographic diversity in ImageNet was lacking \cite{deng2009imagenet}.

First, candidate images for each synset were sourced from commercial image search engines, including Google, Yahoo!, Microsoft's Live Search, Picsearch and Flickr \cite{fei2010imagenet}. Gender \cite{kay2015unequal} and racial \cite{noble2018algorithms} biases have been demonstrated to exist in image search results (i.e. images of occupations), demonstrating that a more curated approach at the top of the funnel may be necessary to mitigate inherent biases of search engines. Second, English search queries were translated into Chinese, Spanish, Dutch and Italian using WordNet databases and used for image retrieval. While this is a step in the right direction, Chinese was the only non-Western European language used, and there exists, for example, Universal Multilingual WordNet which includes over 200 languages for translation \cite{de2009towards}. Third, the authors quantify image diversity by computing the average image of each synset and measuring the lossless JPG file size. They state that a diverse synset will result in a blurrier average image and smaller file, representative of diversity in appearance, position, viewpoint and background. This method, however, cannot quantify diversity with respect to demographic characteristics such as age, gender, and skin type.

\begin{table}
\begin{center}
\footnotesize
\begin{tabular}{|
>{}c |
>{}c |
>{}c |
>{}c |
>{}c |}
\hline
\multicolumn{2}{|c|}{{\color[HTML]{000000} }} & \multicolumn{3}{c|}{{\color[HTML]{000000} \textbf{Gender}}} \\ \cline{3-5} 
\multicolumn{2}{|c|}{\multirow{-2}{*}{{\color[HTML]{000000} \textbf{Accuracy (\%)}}}} & {\color[HTML]{000000} \textbf{Male}} & {\color[HTML]{000000} \textbf{Female}} & {\color[HTML]{000000} \textbf{All}} \\ \hline
{\color[HTML]{000000} } & {\color[HTML]{000000} \textbf{0-14}} & {\color[HTML]{000000} 79.19} & {\color[HTML]{000000} 75.78} & {\color[HTML]{000000} 77.62} \\ \cline{2-5} 
{\color[HTML]{000000} } & {\color[HTML]{000000} \textbf{15-29}} & {\color[HTML]{000000} 96.09} & {\color[HTML]{000000} 89.80} & {\color[HTML]{000000} 91.95} \\ \cline{2-5} 
{\color[HTML]{000000} } & {\color[HTML]{000000} \textbf{30-44}} & {\color[HTML]{000000} 100.00} & {\color[HTML]{000000} 91.72} & {\color[HTML]{000000} 95.99} \\ \cline{2-5} 
{\color[HTML]{000000} } & {\color[HTML]{000000} \textbf{45-59}} & {\color[HTML]{000000} 100.00} & {\color[HTML]{000000} 91.30} & {\color[HTML]{000000} 96.89} \\ \cline{2-5} 
{\color[HTML]{000000} } & {\color[HTML]{000000} \textbf{60+}} & {\color[HTML]{000000} 84.91} & {\color[HTML]{000000} 79.71} & {\color[HTML]{000000} 82.86} \\ \cline{2-5} 
\multirow{-7}{*}{{\color[HTML]{000000} \textbf{\begin{tabular}[c]{@{}c@{}}Age\\ Group\end{tabular}}}} & {\color[HTML]{000000} \textbf{All}} & {\color[HTML]{000000} 94.15} & {\color[HTML]{000000} 88.04} & {\color[HTML]{000000} 91.00} \\ \hline
\end{tabular}
\end{center}
\caption{Gender model binary classification accuracy on APPA-REAL test set}
\label{table:appa-real-gender}
\end{table}

\begin{table}
\begin{center}
\footnotesize
\begin{tabular}{|
>{}c |
>{}c |
>{}c |
>{}c |
>{}c |}
\hline
\multicolumn{2}{|c|}{{\color[HTML]{000000} }} & \multicolumn{3}{c|}{{\color[HTML]{000000} \textbf{Gender}}} \\ \cline{3-5} 
\multicolumn{2}{|c|}{\multirow{-2}{*}{{\color[HTML]{000000} \textbf{Accuracy (\%)}}}} & {\color[HTML]{000000} \textbf{Male}} & {\color[HTML]{000000} \textbf{Female}} & {\color[HTML]{000000} \textbf{All}} \\ \hline
{\color[HTML]{000000} } & {\color[HTML]{000000} \textbf{I}} & {\color[HTML]{000000} 100.00} & {\color[HTML]{000000} 98.31} & {\color[HTML]{000000} 99.12} \\ \cline{2-5} 
{\color[HTML]{000000} } & {\color[HTML]{000000} \textbf{II}} & {\color[HTML]{000000} 100.00} & {\color[HTML]{000000} 97.14} & {\color[HTML]{000000} 98.86} \\ \cline{2-5} 
{\color[HTML]{000000} } & {\color[HTML]{000000} \textbf{III}} & {\color[HTML]{000000} 100.00} & {\color[HTML]{000000} 95.08} & {\color[HTML]{000000} 97.67} \\ \cline{2-5} 
{\color[HTML]{000000} } & {\color[HTML]{000000} \textbf{IV}} & {\color[HTML]{000000} 100.00} & {\color[HTML]{000000} 86.11} & {\color[HTML]{000000} 92.31} \\ \cline{2-5} 
{\color[HTML]{000000} } & {\color[HTML]{000000} \textbf{V}} & {\color[HTML]{000000} 100.00} & {\color[HTML]{000000} 68.79} & {\color[HTML]{000000} 83.70} \\ \cline{2-5} 
{\color[HTML]{000000} } & {\color[HTML]{000000} \textbf{VI}} & {\color[HTML]{000000} 100.00} & {\color[HTML]{000000} 62.77} & {\color[HTML]{000000} 86.22} \\ \cline{2-5} 
\multirow{-7}{*}{{\color[HTML]{000000} \textbf{\begin{tabular}[c]{@{}c@{}}Fitzpatrick\\ Skin\\ Type\end{tabular}}}} & {\color[HTML]{000000} \textbf{All}} & {\color[HTML]{000000} 100.00} & {\color[HTML]{000000} 83.57} & {\color[HTML]{000000} 92.68} \\ \hline
\end{tabular}
\end{center}
\caption{Gender model binary classification accuracy on PPB}
\label{table:ppb-gender}
\end{table}

\section{Methodology}

In order to provide demographic annotations at scale, there exist two feasible methods: crowdsourcing and model-driven annotations. In the case of large-scale image datasets, crowdsourcing quickly becomes prohibitively expensive; ImageNet, for example, employed 49k AMT workers during its collection \cite{liimagenet}. Model-driven annotations use supervised learning methods to create models that can predict annotations, but this approach comes with its own meta-problem; as the goal of this work is to identify demographic representation in data, we must analyze the annotation models for their performance on intersectional groups to determine if they themselves exhibit bias. 

\subsection{Face Detection}

The FaceBoxes network \cite{zhang2017faceboxes} is employed for face detection, consisting of a lightweight CNN that incorporates novel Rapidly Digested and Multiple Scale Convolutional Layers for speed and accuracy, respectively. This model was trained on the WIDER FACE dataset \cite{yang2016wider} and achieves average precision of 95.50\% on the Face Detection Data Set and Benchmark (FDDB) \cite{fddbTech}. On a subset of 1,000 images from FDDB hand-annotated by the author for apparent age and gender, the model achieves a relative fair performance across intersectional groups, as show in Table \ref{table:fddb-face-detection}.

\subsection{Apparent Age Annotation}

The task of apparent age annotation arises as ground-truth ages of individuals in images are not possible to obtain in the domain of web-scraped datasets. In this work, we follow Merler \textit{et al.} \cite{merler2019diversity} and employ the Deep EXpectation (DEX) model of apparent age \cite{rothe2018deep}, which is pre-trained on the IMDB-WIKI dataset of 500k faces with real ages and fine-tuned on the APPA-REAL training and validation sets of 3.6k faces with apparent ages, crowdsourced from an average of 38 votes per image \cite{agustsson2017apparent}. As show in Table \ref{table:appa-real-age}, the model achieves a mean average error of 5.22 years on the APPA-REAL test set, but exhibits worse performance on younger and older age groups.

\begin{table}
\begin{center}
\footnotesize
\begin{tabular}{|
>{}c |
>{}c |
>{}c |
>{}c |
>{}c |}
\hline
\multicolumn{2}{|c|}{{\color[HTML]{000000} }} & \multicolumn{3}{c|}{{\color[HTML]{000000} \textbf{Gender}}} \\ \cline{3-5} 
\multicolumn{2}{|c|}{\multirow{-2}{*}{{\color[HTML]{000000} \textbf{\% of Dataset}}}} & {\color[HTML]{000000} \textbf{Male}} & {\color[HTML]{000000} \textbf{Female}} & {\color[HTML]{000000} \textbf{All}} \\ \hline
{\color[HTML]{000000} } & {\color[HTML]{000000} \textbf{0-14}} & {\color[HTML]{000000} 4.06} & {\color[HTML]{000000} 3.01} & {\color[HTML]{000000} 7.08} \\ \cline{2-5} 
{\color[HTML]{000000} } & {\color[HTML]{000000} \textbf{15-29}} & {\color[HTML]{000000} 27.11} & {\color[HTML]{000000} 23.72} & {\color[HTML]{000000} \textbf{50.83}} \\ \cline{2-5} 
{\color[HTML]{000000} } & {\color[HTML]{000000} \textbf{30-44}} & {\color[HTML]{000000} 17.81} & {\color[HTML]{000000} 9.02} & {\color[HTML]{000000} 26.83} \\ \cline{2-5} 
{\color[HTML]{000000} } & {\color[HTML]{000000} \textbf{45-59}} & {\color[HTML]{000000} 8.48} & {\color[HTML]{000000} 5.08} & {\color[HTML]{000000} 13.56} \\ \cline{2-5} 
{\color[HTML]{000000} } & {\color[HTML]{000000} \textbf{60+}} & {\color[HTML]{000000} 0.91} & {\color[HTML]{000000} 0.80} & {\color[HTML]{000000} 1.71} \\ \cline{2-5} 
\multirow{-7}{*}{{\color[HTML]{000000} \textbf{\begin{tabular}[c]{@{}c@{}}Age\\ Group\end{tabular}}}} & {\color[HTML]{000000} \textbf{All}} & {\color[HTML]{000000} \textbf{58.38}} & {\color[HTML]{000000} 41.62} & {\color[HTML]{000000} 100.00} \\ \hline
\end{tabular}
\end{center}
\caption{Top-level statistics of ILSVRC 2012 ImageNet subset}
\label{table:ILSVRC-stats}
\end{table}

\begin{table}
\begin{center}
\footnotesize
\begin{tabular}{|c|c|c|c|}
\hline
\multicolumn{2}{|c|}{\textbf{\% of Synset Annotated Male}} & \multicolumn{2}{c|}{\textbf{\% of Synset Annotated Female}} \\ \hline
`\textit{barracouta}' & 94.24 & `\textit{brassiere}' & 88.59 \\ \hline
`\textit{ballplayer}' & 93.47 & `\textit{golden retriever}' & 88.29 \\ \hline
`\textit{Windsor tie}' & 93.29 & `\textit{bikini}' & 85.91 \\ \hline
`\textit{gar}' & 92.76 & `\textit{maillot}' & 85.90 \\ \hline
`\textit{tench}' & 92.40 & `\textit{tank suit}' & 85.47 \\ \hline
`\textit{rugby ball}' & 91.96 & `\textit{miniskirt}' & 85.14 \\ \hline
`\textit{barbershop}' & 90.06 & `\textit{cocker spaniel}' & 83.82 \\ \hline
`\textit{bulletproof vest}' & 89.33 & `\textit{gown}' & 83.62 \\ \hline
`\textit{sax}' & 89.16 & `\textit{lipstick}' & 82.81 \\ \hline
`\textit{swimming trunks}' & 88.36 & `\textit{Maltese dog}' & 82.06 \\ \hline
`\textit{assault rifle}' & 88.12 & `\textit{stole}' & 81.64 \\ \hline
`\textit{sturgeon}' & 87.61 & `\textit{cardigan}' & 81.32 \\ \hline
\end{tabular}
\end{center}
\caption{Gender-biased synsets, ILSVRC 2012 ImageNet subset}
\label{table:ILSVRC-synsets}
\end{table}

\subsection{Gender Annotation}

We recognize that a binary representation of gender does not adequately capture the complexities of gender or represent transgender identities. In this work, we express gender as a continuous value between 0 and 1. When thresholding at 0.5, we use the sex labels of `\textit{male}' and `\textit{female}' to define gender classes, as training datasets and evaluation benchmarks use this binary label system. We again follow Merler \textit{et al.} \cite{merler2019diversity} and employ a DEX model to annotate the gender of an individual. When tested on APPA-REAL, with enhanced annotations provided by \cite{clapes2018apparent}, the model achieves an accuracy of 91.00\%, however its errors are not evenly distributed, as shown in Table \ref{table:appa-real-gender}. The model errs more on younger and older age groups and on those with a female gender label. 

Given these biased results, we further evaluate the model on the Pilot Parliaments Benchmark (PPB) \cite{buolamwini2018gender}, a face dataset developed by Buolamwini and Gebru for parity in gender and skin type. Results for intersectional groups on PPB are shown in Table \ref{table:ppb-gender}. The model performs very poorly for darker-skinned females (Fitzpatrick skin types IV - VI), with an average accuracy of 69.00\%, reflecting the disparate findings of commercial computer vision gender classifiers in \textit{Gender Shades} \cite{buolamwini2018gender}. We note that use of this model in annotating ImageNet will result in biased gender annotations, but proceed in order to establish a baseline upon which the results of a more fair gender annotation model can be compared in future work, via fine-tuning on crowdsourced gender annotations from the \textit{Diversity in Faces} dataset \cite{merler2019diversity}.

\begin{table}
\begin{center}
\footnotesize
\begin{tabular}{|
>{}c |
>{}c |
>{}c |
>{}c |
>{}c |}
\hline
\multicolumn{2}{|c|}{{\color[HTML]{000000} }} & \multicolumn{3}{c|}{{\color[HTML]{000000} \textbf{Gender}}} \\ \cline{3-5} 
\multicolumn{2}{|c|}{\multirow{-2}{*}{{\color[HTML]{000000} \textbf{\% of Dataset}}}} & {\color[HTML]{000000} \textbf{Male}} & {\color[HTML]{000000} \textbf{Female}} & {\color[HTML]{000000} \textbf{All}} \\ \hline
{\color[HTML]{000000} } & {\color[HTML]{000000} \textbf{0-14}} & {\color[HTML]{000000} 3.35} & {\color[HTML]{000000} 1.66} & {\color[HTML]{000000} 5.00} \\ \cline{2-5} 
{\color[HTML]{000000} } & {\color[HTML]{000000} \textbf{15-29}} & {\color[HTML]{000000} 24.63} & {\color[HTML]{000000} 16.95} & {\color[HTML]{000000} \textbf{41.58}} \\ \cline{2-5} 
{\color[HTML]{000000} } & {\color[HTML]{000000} \textbf{30-44}} & {\color[HTML]{000000} 21.02} & {\color[HTML]{000000} 7.36} & {\color[HTML]{000000} 28.38} \\ \cline{2-5} 
{\color[HTML]{000000} } & {\color[HTML]{000000} \textbf{45-59}} & {\color[HTML]{000000} 16.87} & {\color[HTML]{000000} 3.70} & {\color[HTML]{000000} 20.57} \\ \cline{2-5} 
{\color[HTML]{000000} } & {\color[HTML]{000000} \textbf{60+}} & {\color[HTML]{000000} 3.02} & {\color[HTML]{000000} 1.44} & {\color[HTML]{000000} 4.47} \\ \cline{2-5} 
\multirow{-7}{*}{{\color[HTML]{000000} \textbf{\begin{tabular}[c]{@{}c@{}}Age\\ Group\end{tabular}}}} & {\color[HTML]{000000} \textbf{All}} & {\color[HTML]{000000} \textbf{68.89}} & {\color[HTML]{000000} 31.11} & {\color[HTML]{000000} 100.00} \\ \hline
\end{tabular}
\end{center}
\caption{Top-level statistics of ImageNet `\textit{person}' subset}
\label{table:Person-stats}
\end{table}

\begin{table}
\begin{center}
\footnotesize
\begin{tabular}{|c|c|c|c|}
\hline
\multicolumn{2}{|c|}{\textbf{\% of Synset Annotated Male}} & \multicolumn{2}{c|}{\textbf{\% of Synset Annotated Female}} \\ \hline
`\textit{spree}' & 100.00 & `\textit{bombshell}' & 100.00 \\ \hline
`\textit{pitchman}' & 100.00 & `\textit{choker}' & 100.00 \\ \hline
`\textit{counterterrorist}' & 100.00 & `\textit{dyspeptic}' & 100.00 \\ \hline
`\textit{second}' & 100.00 & `\textit{comedienne}' & 96.30 \\ \hline
`\textit{Kennan}' & 100.00 & `\textit{cover}' & 96.24 \\ \hline
`\textit{chandler}' & 100.00 & `\textit{tempter}' & 95.92 \\ \hline
`\textit{agnostic}' & 100.00 & `\textit{nymph}' & 95.74 \\ \hline
`\textit{helmsman}' & 100.00 & `\textit{artist}''s & 95.05 \\ \hline
`\textit{anti-American}' & 100.00 & `\textit{nymphet}' & 93.84 \\ \hline
`\textit{Girondist}' & 100.00 & `\textit{smasher}' & 93.84 \\ \hline
`\textit{argonaut}' & 100.00 & `\textit{maid}' & 93.05 \\ \hline
`\textit{oilman}' & 100.00 & `\textit{model}' & 93.01 \\ \hline
\end{tabular}
\end{center}
\caption{Top-level statistics of ImageNet `\textit{person}' subset}
\label{table:Person-synsets}
\end{table}

\section{Results}

We evaluate the training set of the ILSVRC 2012 subset of ImageNet (1000 synsets) and the `\textit{person}' hierarchical synset of ImageNet (2833 synsets) with the proposed methodology. Face detections that receive a confidence score of 0.9 or higher move forward to the annotation phase. Statistics for both datasets are presented in Tables \ref{table:ILSVRC-stats} and \ref{table:Person-stats}. In these preliminary annotations, we find that females comprise only 41.62\% of images in ILSVRC and 31.11\% in the `\textit{person}' subset of ImageNet, and people over the age of 60 are almost non-existent in ILSVRC, accounting for 1.71\%.

To get a sense of the most biased classes in terms of gender representation for each dataset, we filter synsets that contain at least 20 images in their class and received face detections for at least 15\% of their images. We then calculate the percentage of males and females in each synset and rank them in descending order. Top synsets for each gender and dataset are presented in Tables \ref{table:ILSVRC-synsets} and \ref{table:Person-synsets}. Top ILSVRC synsets for males largely represent types of fish, sports and firearm-related items and top synsets for females largely represent types of clothing and dogs.

\section{Conclusion}

Through the introduction of a preliminary pipeline for automated demographic annotations, this work hopes to provide insight into the ImageNet dataset, a tool that is commonly abstracted away by the computer vision community. In the future, we will continue this work to create fair models for automated demographic annotations, with emphasis on the gender annotation model. We aim to incorporate additional measures of diversity into the pipeline, such as Fitzpatrick skin type and other craniofacial measurements. When annotation models are evaluated as fair, we plan to continue this audit on all 14.2M images of ImageNet and other large image datasets. With accurate coverage of the demographic attributes of ImageNet, we will be able to investigate the downstream impact of under- and over-represented groups in the features learned in pretrained CNNs and how bias represented in these features may propagate in transfer learning to new applications.

{\small
\bibliographystyle{ieee}
\bibliography{main}
}

\end{document}